%% file: main.tex
\documentclass[10pt,twocolumn,letterpaper]{article}
\usepackage{cvpr}
\input{preamble}
\definecolor{cvprblue}{rgb}{0.21,0.49,0.74}
\usepackage[pagebackref,breaklinks,colorlinks,citecolor=cvprblue]{hyperref}

\definecolor{gr}{rgb}{0.906, 0.902, 0.902}
\definecolor{bl}{rgb}{0.851, 0.882, 0.949}
\definecolor{dg}{rgb}{0, 0.5, 0}
\def\paperID{9086} 
\def\confName{CVPR}
\def\confYear{2024}

\title{Improving Visual Recognition with Hyperbolical Visual Hierarchy Mapping}
\author{Hyeongjun Kwon$^1$ \quad Jinhyun Jang$^1$\quad Jin Kim$^1$ \quad Kwonyoung Kim$^1$ \quad Kwanghoon Sohn$^{1, 2}$\thanks{Corresponding author.}\\
$^1$Yonsei University, \quad  $^2$Korea Institute of Science and Technology (KIST)\\
{\tt\small \{kwonjunn01, jr000192, kimjin928, kyk12, khsohn\}@yonsei.ac.kr},}

\begin{document}
\maketitle
 \blfootnote{This research was supported by the National Research Foundation of Korea (NRF) grant funded by the Korea government (MSIP) (NRF2021R1A2C2006703).}
\input{sec/0_abstract}
\vspace{-10pt}
\input{sec/1_intro}

\input{sec/2_related_new}
\input{sec/3_prelim}
\input{sec/4_method}

\input{sec/5_experiment}
\input{sec/6_conclusion}

{
\small
\bibliography{main}
}
\clearpage
\input{supplementary}
\end{document}

%% file: preamble.tex
\usepackage[dvipsnames]{xcolor}
\usepackage{graphicx}
\usepackage{amsmath}
\usepackage{amssymb}
\usepackage[linesnumbered,ruled,vlined]{algorithm2e}
\usepackage{tabularx,booktabs}
\usepackage{color}
\usepackage{colortbl}
\usepackage{multirow}
\usepackage{multicol}
\usepackage{subcaption}
\usepackage{tabularray}
\UseTblrLibrary{diagbox}
\usepackage{pifont}
\usepackage{lipsum}  
\usepackage{makecell}
\usepackage{subcaption}
\usepackage{babel}
\usepackage{microtype}
\newcommand{\red}[1]{{\color{red}#1}}
\newcommand{\blue}[1]{{\color{blue}#1}}

\newcommand{\figref}[1]{Fig.~\ref{#1}}
\newcommand{\tabref}[1]{Tab.~\ref{#1}}
\newcommand{\equref}[1]{Eq.~(\ref{#1})}
\newcommand{\secref}[1]{Sec.~\ref{#1}}

\newcommand{\cmark}{\ding{51}}%
\newcommand{\xmark}{\ding{55}}%

\usepackage[accsupp]{axessibility}
\definecolor{LinkWater}{rgb}{0.85,0.882,0.949}
\newcommand\blfootnote[1]{%
  \begingroup
  \renewcommand\thefootnote{}\footnote{#1}%
  \addtocounter{footnote}{-1}%
  \endgroup
}     

%% file: sec/0_abstract.tex
\begin{abstract}
Visual scenes are naturally organized in a hierarchy, where a coarse semantic is recursively comprised of several fine details.
Exploring such a visual hierarchy is crucial to recognize the complex relations of visual elements, leading to a comprehensive scene understanding.
In this paper, we propose a Visual Hierarchy Mapper (Hi-Mapper), a novel approach for enhancing the structured understanding of the pre-trained Deep Neural Networks (DNNs).
Hi-Mapper investigates the hierarchical organization of the visual scene by 1) pre-defining a hierarchy tree through the encapsulation of probability densities;
and 2) learning the hierarchical relations in hyperbolic space with a novel hierarchical contrastive loss.
The pre-defined hierarchy tree recursively interacts with the visual features of the pre-trained DNNs through hierarchy decomposition and encoding procedures, thereby effectively identifying the visual hierarchy and enhancing the recognition of an entire scene.
Extensive experiments demonstrate that Hi-Mapper significantly enhances the representation capability of DNNs, leading to an improved performance on various tasks, including image classification and dense prediction tasks.
The code is available at \url{https://github.com/kwonjunn01/Hi-Mapper}.
\end{abstract}

%% file: sec/1_intro.tex
\vspace{-10pt}
\section{Introduction}
\label{sec:intro}
Recognizing and representing the visual scene of any content is the fundamental pursuit of the computer vision field~\cite{hypvit,hier,gkioxari2018detecting,jang2023knowing}.
In particular, understanding \textit{what constitutes a scene} and \textit{how each element is comprised of} plays a key role in various visual recognition tasks such as image retrieval~\cite{hypvit,hier}, human-object interaction~\cite{hou2021affordance,gkioxari2018detecting}, and dense prediction~\cite{kwon2023probabilistic,pin}.
This goes beyond merely learning discriminative feature representations, as it requires to reason about the fine details as well as their associations to comprehend the structured nature of the complex visual scene.
\begin{figure}[!t]
      \begin{subfigure}{0.461\linewidth}
          \includegraphics[width=1\linewidth]{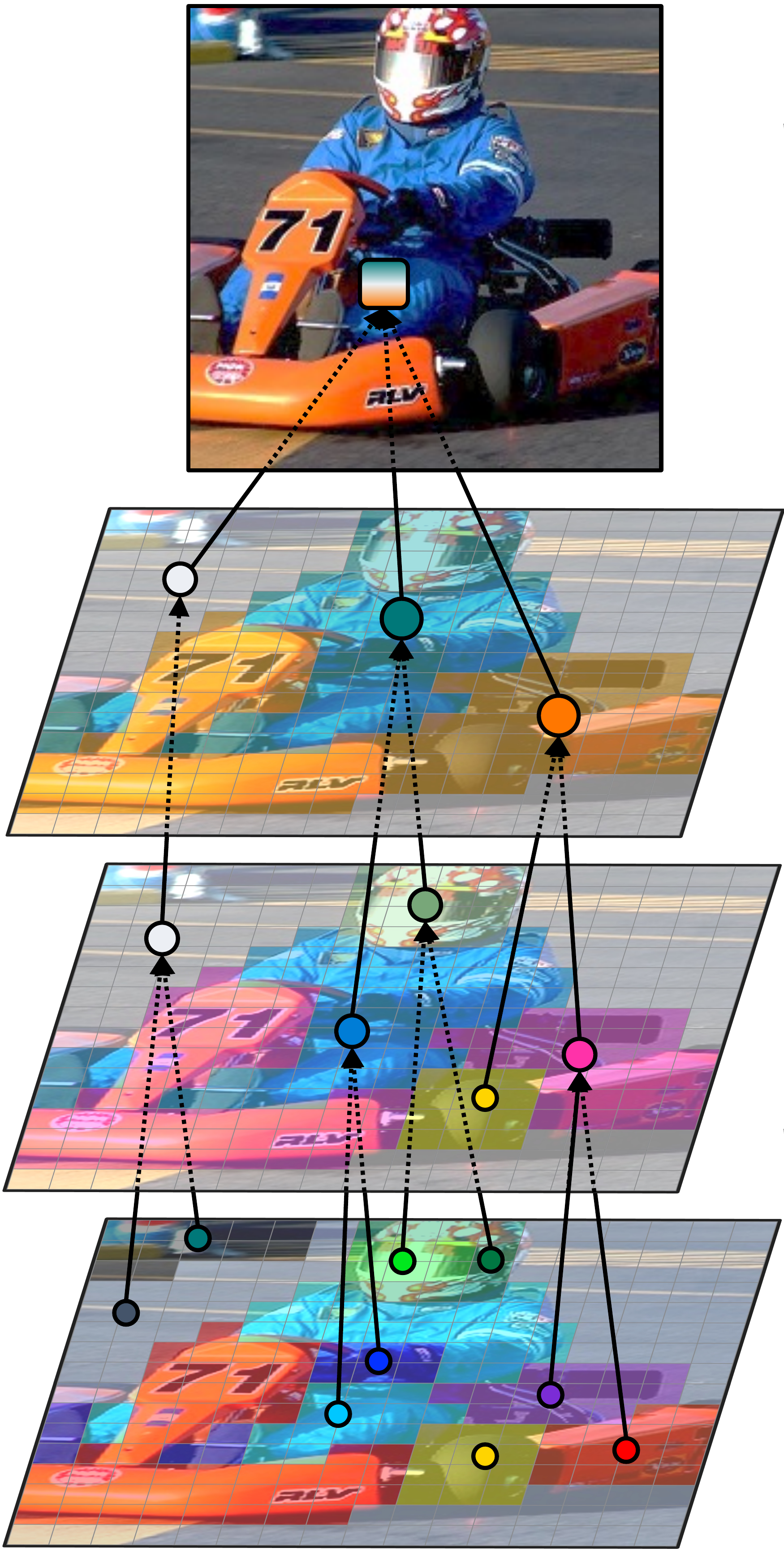}
          \caption{}
          \label{fig:1a}
      \end{subfigure}
      \begin{subfigure}{0.53\linewidth}
          \includegraphics[width=1\linewidth]{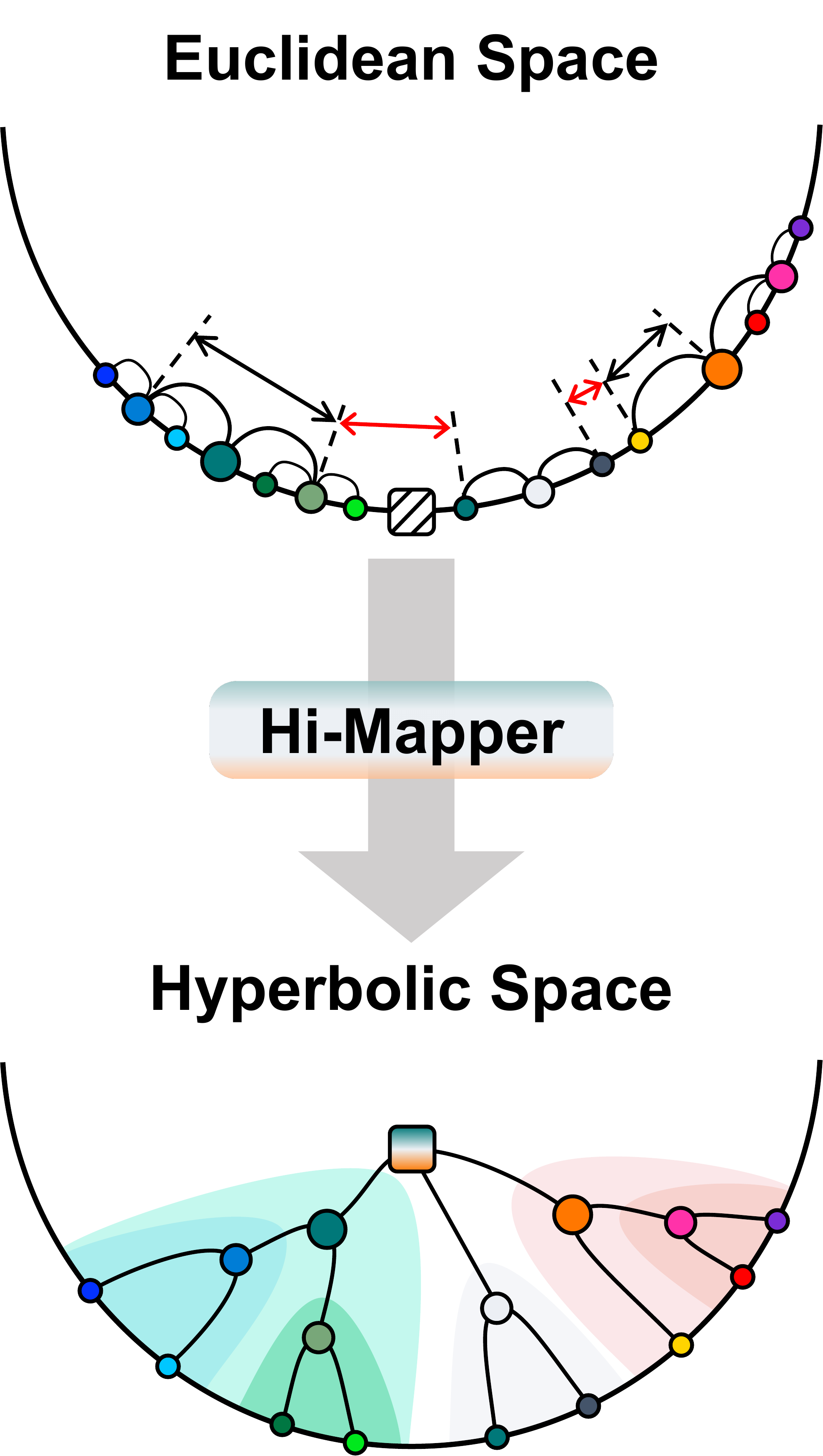}
          \vspace{-15pt}
          \caption{}
          \label{fig:1b}
      \end{subfigure}
    \vspace{-15pt}
    \caption{(a) A visual scene can be decomposed into a hierarchical structure based on the semantics of each visual element.
    (b) Euclidean space is suboptimal in representing the hierarchical structure due to its flat nature.
    The relational distance is inaccurately captured, being unaware of the semantic similarity of visual elements (Red line).
    Hi-Mapper maps the hierarchical elements in hyperbolic space, which effectively preserves their semantic relations and distances due to its constant negative curvature.
    }
\vspace{-15pt}
\label{fig:concept}
\end{figure}

Over the decades, the development of deep neural networks (DNNs) has contributed towards advances in representing the complex visual scene.
Notably, convolutional neural networks (CNNs) have achieved capturing fine details through the local convolutional filters while Vision Transformer (ViT)~\cite{vit} has enabled coarse context modeling with multi-head self-attention mechanisms. 
Owing to their different desirable properties, hybrid architectures~\cite{ramachandran2019stand,zhao2020exploring,cmt,cvt} and multi-scale variants of ViTs~\cite{dong2022cswin,wang2021pyramid,mvit2} have been extensively explored to capitalize on the complementary features of CNNs and ViTs.
Subsequent works~\cite{crossvit,dynawindow} have further imposed interaction between multi-scale image patches to facilitate information exchange between fine details and coarse semantics.

While they have shown effective in capturing coarse-to-fine information, a structured understanding of the visual scene remains underexplored.
Concretely, a scene can be interpreted as a hierarchical composition of visual elements, where the ability to recognize an instance (\eg a man) at a coarse level arises from the ability to compose its constituents (\eg body parts) at a finer level, as shown in \figref{fig:1a}.
Being aware of such semantic hierarchies furthers the perception of grid-like geometry and enhances the recognition of an entire image.
Drawn from the same motivation, the recent variants of ViT~\cite{quadtree,ding2023visual,ke2023learning} have constructed the hierarchies between image tokens across the transformer layers.
However, their hierarchical relations are defined through the symmetric measurement (\ie, cosine similarity) which lacks the ability to represent the asymmetric property of hierarchical structure (\ie, inclusion of parent-child nodes). 
In addition, their image tokens are represented in Euclidean space where the hierarchical relations become distorted due to its linear and flat geometry~\cite{365733,distort,hyp1s,hyplorentz,hypfs, hyptxt1s, hyptxt2s,hypgraph1s,hypmeru}, as shown in \figref{fig:1b}.

In this paper, we propose a novel Visual Hierarchy Mapper (Hi-Mapper) that improves the structured understanding of pre-trained DNNs by identifying the visual hierarchy.
Hi-Mapper accomplishes this through:
1) Probabilistic modeling of hierarchy nodes, where the mean vector and covariance represent the center and scale of visual-semantic cluster, respectively~\cite{gaussian1,gaussian2,gaussian3,biasrelation3}.
Accordingly, the asymmetric hierarchical relations are captured through the inclusion of probability densities.
Furthermore, 2) Hi-Mapper maps the hierarchy nodes to hyperbolic space, where its constant negative curvature effectively represents the exponential growth of hierarchy nodes.
Specifically, Hi-Mapper pre-defines a tree-like structure, with its leaf-level node modeled as a unique Gaussian distribution and the higher-level nodes approximated by a Mixture of Gaussians (MoG) of their corresponding child nodes.
The pre-defined hierarchy nodes then interact with the penultimate visual feature map of the pre-trained DNNs to decompose the feature map into the visual hierarchy.
Moreover, in order to bypass the difficulties of modeling hierarchy in Euclidean space, Hi-Mapper maps the identified visual hierarchy to hyperbolic space and learns the hierarchical relation with a novel hierarchical contrastive loss.
The proposed loss enforces the child-parent nodes to be similar and child-child nodes to be dissimilar in a shared hyperbolic space.
The visual hierarchy then interacts with the global visual feature of the pre-trained DNNs such that the hierarchical relations are fully encoded in the global feature representation.

Hi-Mapper serves as a plug-and-play module, which generalizes over any type of DNNs and flexibly identifies the hierarchical organization of visual scenes.
We conduct extensive experiments with various pre-trained DNNs (\ie, ResNet~\cite{resnet}, DeiT~\cite{deit}) on several benchmarks~\cite{deng2009imagenet, lin2014microsoft, zhou2017scene} to demonstrate the effectiveness of Hi-Mapper.

In summary, our key contributions are as follows:
\begin{itemize}
\item We present a novel Visual Hierarchy Mapper (Hi-Mapper) that enhances the structured understanding of the pre-trained DNNs by investigating the hierarchical organization of visual scene.
The proposed Hi-Mapper is applicable to any type of the pre-trained DNNs without modifying the underlying structures.
\item Hi-Mapper effectively identifies the visual hierarchy by combining the favorable characteristic of probabilistic modeling and hyperbolic geometry for representing the hierarchical structure.
\item We conduct extensive experiments to validate the efficacy of the proposed Hi-Mapper, and improves over the state-of-the-art approaches on various visual recognition tasks.
\end{itemize}

%% file: sec/2_related_new.tex
\section{Related Work}
\label{sec:related}
\paragraph{Hierarchy-aware visual recognition.}
Unsupervised image parsing is a long-standing pursuit in visual recognition tasks from the classical computer vision era~\cite{classicalparsing1,classicalparsing2,classicalparsing3,classicalparsing4,classicalparsing5}.
In the pre-deep learning era, Zhouwen~\etal~\cite{classicalparsing4} firstly introduces a framework to parse images with their constituents via a divide-and-conquer strategy.
CapsuleNet-based methods have demonstrated substantial enhancements in image parsing, facilitated by dynamic routing, which efficiently capture the compositional relationships among the activities of capsules that represent object parts.
Recently, hierarchical semantic segmentation has been extensively researched, including human parser~\cite{humabparse1,humabparse2} based on human-part hierarchy and unsupervised part segmentations~\cite{unsuppartseg1,unsuppartseg2,unsuppartseg3}.

Beyond image parsing, recent researches on deep neural networks (DNNs) have attempted to exploit hierarchical relationships between detail and global representations.
CrossViT~\cite{crossvit} utilizes a dual-branch transformer for multi-scale feature extraction, enriching features through a fusion module that integrates inter-scale patch relationships.
Quadtree~\cite{quadtree} iteratively and hierarchically selects a subset of crucial finer patches within each coarse patch.
DependencyViT~\cite{ding2023visual} inverts the self-attention process to organize patches as parent and child nodes, enabling a hierarchical exploration.
More recently, CAST~\cite{cast} employs superpixel-based patch generation and graph pooling for hierarchical patch merging for improving fine-grained recognition performances.
While they define hierarchical relations with token similarities in Euclidean space, we pre-define a hierarchical structure with probabilistic modeling and learn the relation in hyperbolic space.

\vspace{-10pt}
\paragraph{Probabilistic modeling.}
Probabilistic representation has been extensively explored in the early NLP studies for handling the nuance of word semantics with the probability distribution.
For instance, Vilnis \etal~\cite{gaussian1} first introduced the probability densities for representing word embeddings.
Athiwaratkun \etal~\cite{gaussian2} discovered that an imbalance in word frequency leads to distortions in word order and mitigated the problem by representing the word orders through the encapsulation of probability densities.
Besides word representation, abundant research has demonstrated the effectiveness of probabilistic modeling in visual representation~\cite{chun2021probabilistic,kwon2023probabilistic,shi2019probabilistic,park2022probabilistic}.
For example, Shi~\etal~\cite{shi2019probabilistic} proposed to penalize the low quality face images by measuring the variance of each image distribution.
Chun \etal~\cite{chun2021probabilistic} identified the limitations of deterministic modeling in vision-language domains and introduced probabilistic cross-modal embedding for providing the uncertainty estimates.

In this work, we deploy probabilistic modeling in defining hierarchical structure, where each distribution represents the inclusive relations of hierarchy nodes.

\vspace{-10pt}
\paragraph{Hyperbolic manifold.}
Hyperbolic manifolds have gained increasing interest in deep learning area due to their effectiveness in modeling hierarchical structures.
Their success in NLP field~\cite{hyptxt1s,hyptxt2s,hyplorentz,hyppoincare}
has inspired approaches to adopt hyperbolic manifolds in computer vision researches such as image retrieval~\cite{hier,hypvit}, image segmentation~\cite{hypseg1,hypseg2}, and few-shot learning~\cite{hypfs}.
As a pioneering work, Khrulkov \etal~\cite{khrulkov2020hyperbolic} investigated an exponential map from Euclidean space to hyperbolic space for learning hierarchical image embeddings.
Ermolov \etal~\cite{hypvit} applied pair-wise cross entropy loss in hyperbolic space for ViTs.
Kim \etal~\cite{hier} extended the work by discovering the latent hierarchy of training data with learnable hierarchical proxies in hyperbolic space.
Focusing on pixel-level analysis, \cite{hypseg2} identified the long-tail objects by embedding masked instance regions into hyperbolic manifolds.
More recently, Desai \etal~\cite{hypmeru} introduced to learn joint image-text embedding space in hyperbolic manifold.
While they explore hyperbolic manifold for representing the categorical hierarchies, we identify the hierarchical structure of visual elements without the part-level annotation through a novel hierarchical contrastive loss.

%% file: sec/3_prelim.tex
\section{Hyperbolic Geometry}
\label{sec:prelim}
{Hyperbolic manifold is a smooth \textit{Riemannian manifold} $\mathcal{M}$ with negative curvature $c$ equipped with a Riemannian metric $g$.
The manifold consists of five isometric models and we utilize the Lorentz model for developing Hi-Mapper due to its training stability.}
A hyperbolic manifold of $n$-dimensions can be represented as a sub-manifold of the Lorentz model $\mathbb{R}^{n+1}$ as an upper half of a two-sheeted hyperboloid.
In the Lorentz space, every point $\mathbf{x}\in\mathbb{R}^{n+1}$ can be denoted as $[\mathbf{x}_\text{space},x_\text{time}]$, where $\mathbf{x}_\text{space}\in\mathbb{R}^n$ and $x_\text{time}\in\mathbb{R}$.
Let $\left\langle\mathbf{x}, \mathbf{y}\right\rangle$ be the Euclidean inner product and $\left\langle\mathbf{x}, \mathbf{y}\right\rangle_\mathbb{L}$ denote the \textit{Lorentzian inner product} which is derived by the Riemannian metric of the Lorentz model $g_\mathbb{L}$.
Given two vectors $\mathbf{x},\mathbf{y}\in\mathbb{R}^{n+1}$, the Lorentzian inner product is computed as follows:
\begin{equation}
    \left\langle\mathbf{x},\mathbf{y}\right\rangle_\mathbb{L}=-x_\text{time}y_\text{time} + \left\langle\mathbf{x}_\text{space}, \mathbf{y}_\text{space}\right\rangle.
\end{equation}
The \(n\)-dimensional Lorentz model $(\mathbb{L}^{n}, g_\mathbb{L})$ is defined by the manifold $\mathbb{L}^{n}=\left\{\mathbf{x}\in\mathbb{R}^{n+1}:\left\langle\mathbf{x},\mathbf{x}\right\rangle_\mathbb{L}=-1/c, c>0\right\}$ and Riemmanian metric of the Lorentz model $g_\mathbb{L}$.
Thus, all vectors satisfy the following constraints:
\begin{equation}
    \label{eq:time}
    -x_\text{time}^{2} + \left\lVert\mathbf{x}_\text{space}\right\rVert^{2} = -1/c.  
\end{equation}
A \textit{geodesic} is the shortest path between two vectors on the manifold. 
The \textit{Lorentzian distance} on $\mathbb{L}$ is then defined as:
\begin{equation}
\label{eq:dist}
    D_\mathbb{L}(\mathbf{x}, \mathbf{y}) = \mathrm{arccosh}(-\left\langle\mathbf{x}, \mathbf{y}\right\rangle_\mathbb{L}).
\end{equation}
The \textit{exponential map} is a way to map vectors from tangent space \(\mathcal{T}_{z}\mathbb{L}^{n}\) onto hyperbolic manifolds \(\mathbb{L}^{n}\), where \(\mathcal{T}_{z}\mathbb{L}^{n}\) is a Euclidean space of vectors that are orthogonal to some point \(\mathbf{z}\in\mathbb{L}^{n}\).
We map the tangent vector $\mathbf{v}\in\mathcal{T}_{z}\mathbb{L}^{n}$ from Euclidean space to the Lorentz manifolds, in which the exponential map $\mathrm{expm}_{\mathbf{z}}$ is defined as: 
\begin{equation}
    \label{eq:expm}
    \mathbf{x} = \mathrm{expm}_{z}(\mathbf{v}) = \cosh({\sqrt{c}\lVert\mathbf{v}\rVert_\mathbb{L}})\mathbf{z}+\frac{\sinh(\sqrt{c}\lVert\mathbf{v}\rVert_\mathbb{L})}{\sqrt{c}\lVert\mathbf{v}\rVert_\mathbb{L}}\mathbf{v}.
\end{equation}
The \textit{logarithm map} $\mathrm{logm}_{z}$ which transfers $\mathbf{x}$ on the hyperboloid back to the tangent space $\mathcal{T}_{z}\mathcal{M}$ is defined as:
\begin{equation}
    \mathbf{v} = \mathrm{logm}_{z}(\mathbf{x}) = \frac{\cosh^{-1}(-c\left\langle\mathbf{z},\mathbf{x}\right\rangle_\mathbb{L})}{\sqrt{(c\left\langle\mathbf{z},\mathbf{x}\right\rangle_\mathbb{L})^{2}-1}}\mathrm{proj}_{\mathbf{z}}(\mathbf{x}).
\end{equation}
We set $\mathbf{z}$ as the origin of the hyperboloid $\mathbf{O} = \left[ \mathbf{0}, \sqrt{1/c}\right]$.

%% file: sec/4_method.tex
\begin{figure*}[!ht]
    \begin{subfigure}[h]{.71\linewidth}
    \centering
    \includegraphics[width=\linewidth]{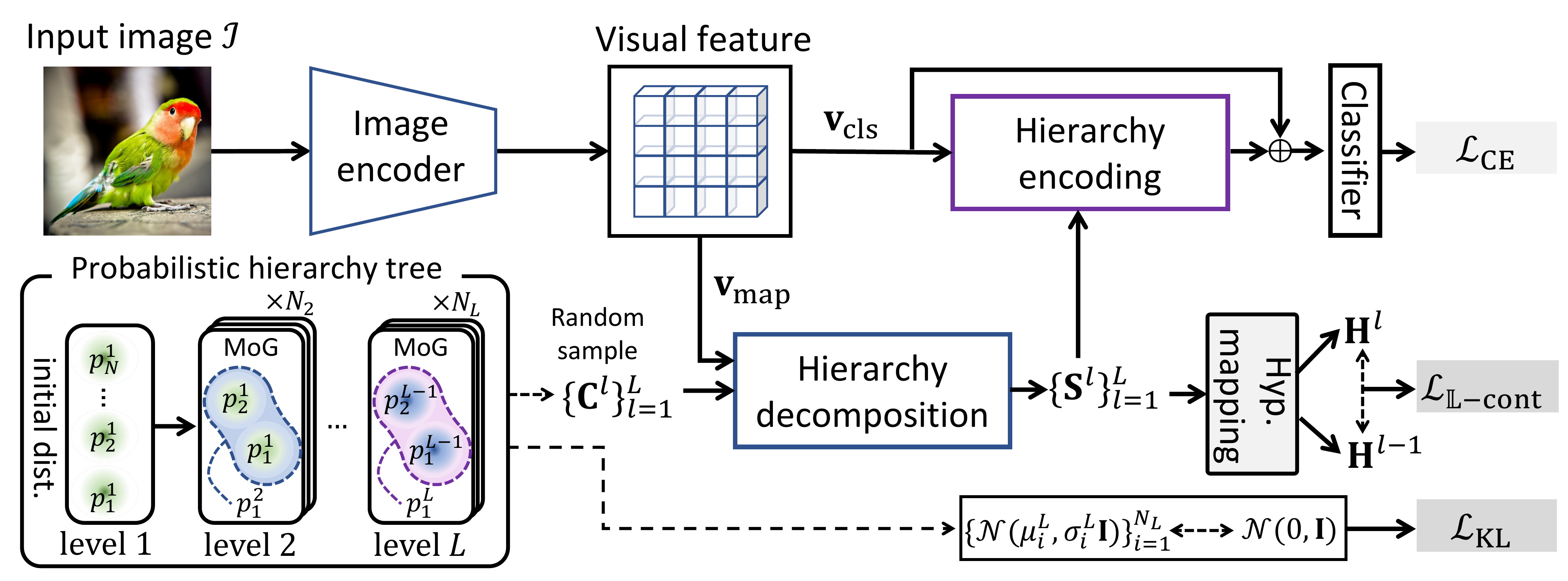}
    \vspace{-4mm}
    \caption{Overall pipeline}
    \label{fig:2a}
    \end{subfigure}
    \begin{subfigure}[h]{.29\linewidth}
    \centering
    \includegraphics[width=\linewidth]{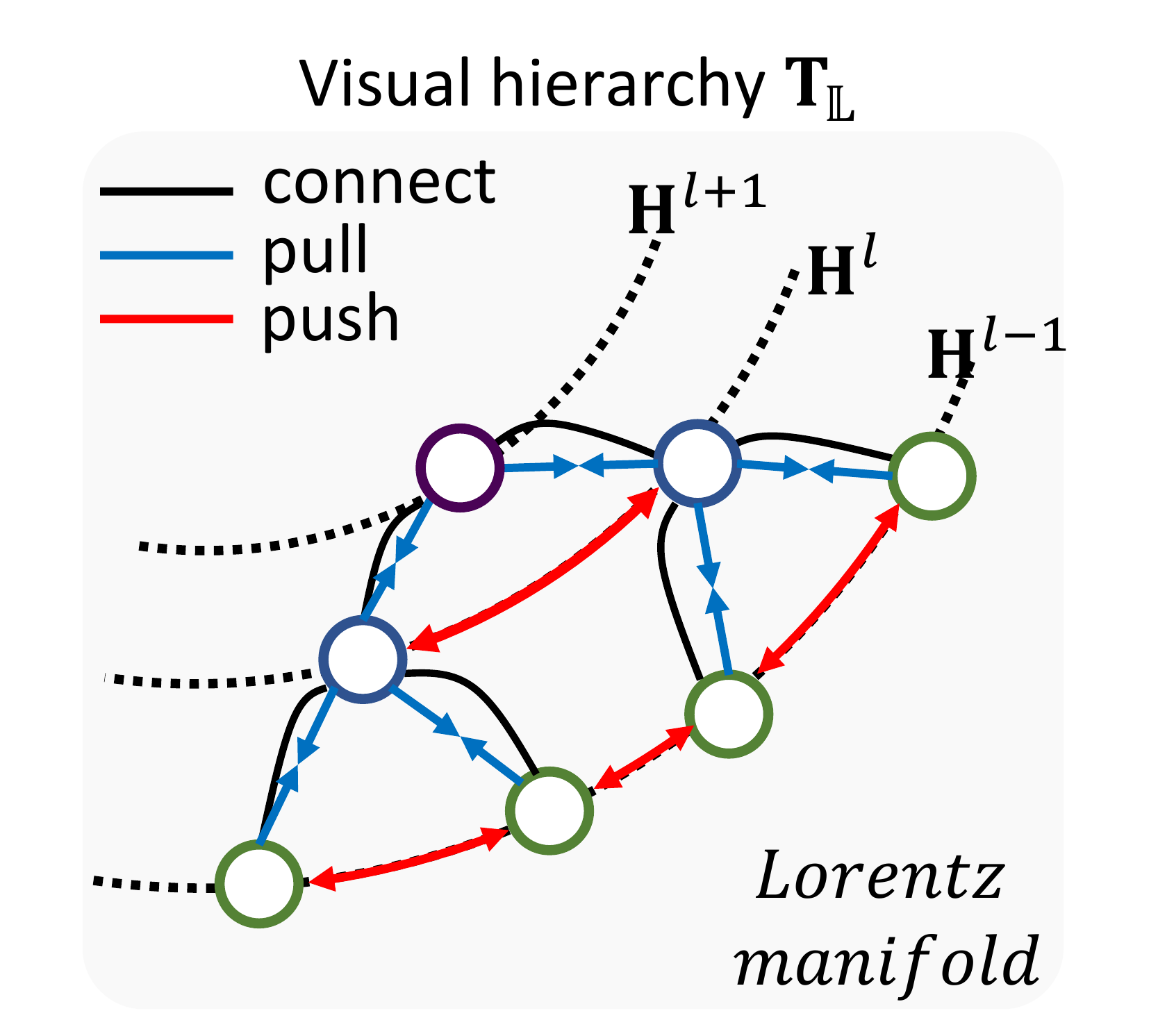}
    \vspace{-4mm}
    \caption{Hierarchical contrastive loss}
    \label{fig:2b}
    \end{subfigure}
    \vspace{-5pt}
    \caption{
(a) An overview of the proposed Hi-Mapper. 
Hi-Mapper operates on top of pre-trained image encoder $\mathcal{F}$, with probabilistic hierarchy tree $\mathbf{T}=\{\mathbf{C}^{l}\}^{L}_{l=1}$.
The tree interacts with visual feature map $\textbf{v}_\text{map}$ through hierarchy decomposition module $\mathcal{D}$, thereby identifying visual hierarchy in Euclidean space $\mathbf{T}_\mathbb{E}=\{\mathbf{S}^{l}\}^{L}_{l=1}$.
The visual hierarchy is mapped to hyperbolic space $\mathbf{T}_{\mathbb{L}} = \{\mathbf{H}^{l}\}_{l=1}^{L}$ and optimized with hierarchical contrastive loss $\mathcal{L}_{\mathbb{L}\text{-cont}}$.
The visual hierarchy is further encoded into global visual representation $\textbf{v}_\text{cls}$ via hierarchy encoding module $\mathcal{G}$ for enhancing the recognition of entire scene.
(b) The proposed hierarchical contrastive loss pulls each parent-child node and pushes all the other nodes at the same level.}
\vspace{-15pt}
\end{figure*}
\begin{figure}[!h]
    \begin{subfigure}[h]{\linewidth}
    \centering
    \includegraphics[width=0.95\linewidth]{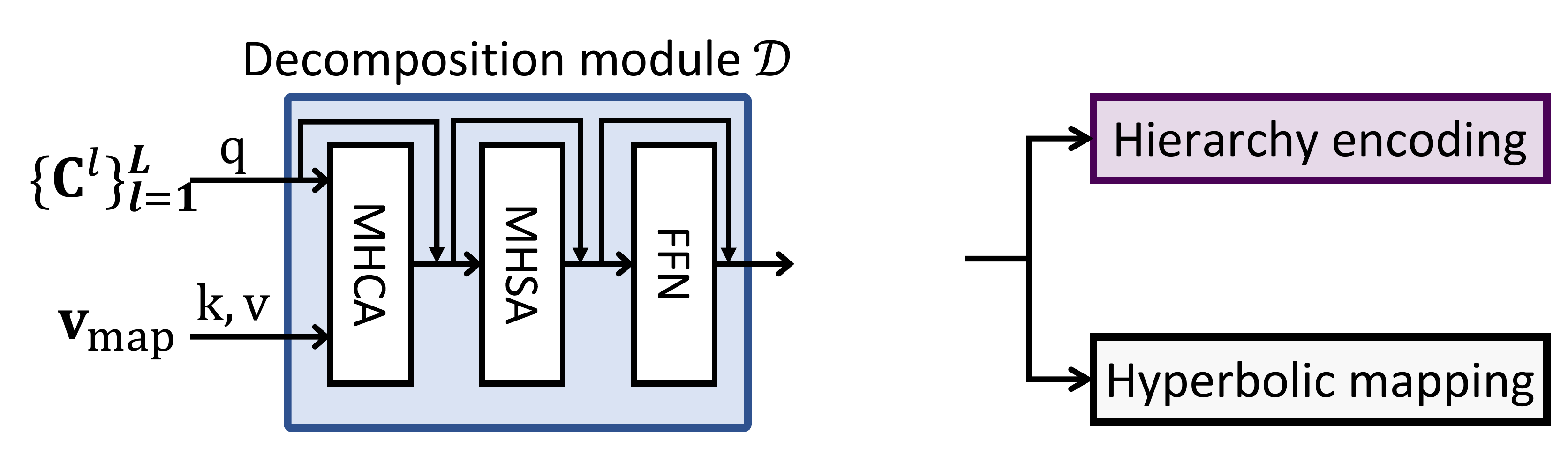}
    \caption{Hierarchy decomposition}
    \label{fig:3a}
    \end{subfigure}
    \begin{subfigure}[h]{\linewidth}
    \centering
    \includegraphics[width=0.95\linewidth]{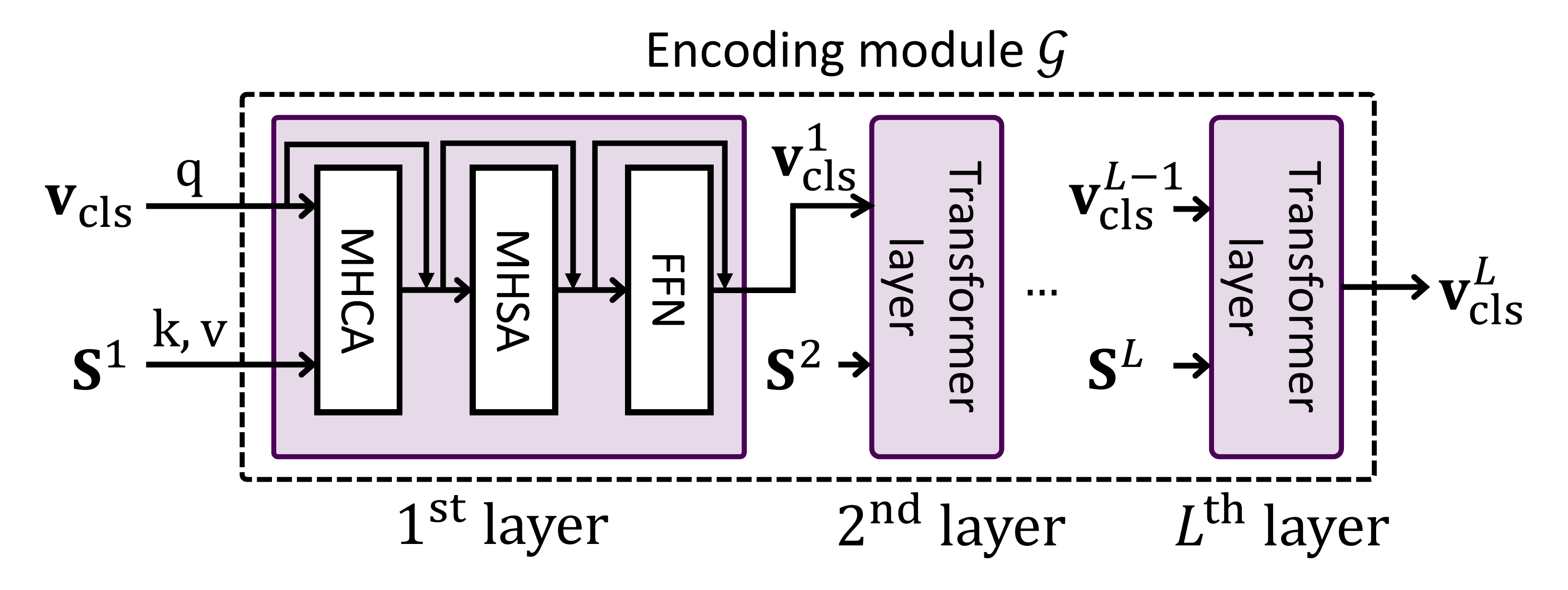}
    \caption{Hierarchy encoding}
    \label{fig:3b}
    \end{subfigure}
    \vspace{-5pt}
    \caption{
    (a) Hierarchy decomposition module groups semantically-relevant visual features \(\mathbf{v}_{\text{map}}\) to the closest semantic cluster \(\mathbf{C}^{l}\).
    (b) Hierarchy encoding module progressively updates global representation \(\mathbf{v}_{\text{cls}}\) by aggregating the visual hierarchy \(\mathbf{S}^{l}\).
    }
\vspace{-10pt}
\label{fig:detail}
\end{figure}
\section{Method} 
\subsection{Overview}~\label{sec:arc&pro}
Our goal is to enhance the structured understanding of pre-trained deep neural networks (DNNs) by investigating the hierarchical organization of visual scenes.
To this end, we introduce a Visual Hierarchy Mapper (Hi-Mapper) which serves as a plug-and-play module on any type of pre-trained DNNs.
An overview of Hi-Mapper is depicted in~\figref{fig:2a}.

Given an image \(\mathcal{I}\), we first extract visual features $[\mathbf{v}_\text{map}, \mathbf{v}_\text{cls}]=\mathcal{F}(\mathcal{I})$ from a pre-trained image encoder \(\mathcal{F}\), where $\mathbf{v}_\text{map}\in\mathbb{R}^{hw \times {d}}$ is the penultimate visual feature map and $\mathbf{v}_\text{cls}\in\mathbb{R}^{d}$ is the global visual representation, with \(hw\) indicating the size of the visual feature map.

Hi-Mapper identifies the visual hierarchy from the visual feature map $\mathbf{v}_\text{map}$ and encodes the identified visual hierarchy back to the global visual representation $\mathbf{v}_\text{cls}$ for enhancing the recognition of a whole scene.
To this end, we pre-define a hierarchy tree with Gaussian distribution, where the relations of the hierarchy nodes are defined through the inclusion of probability densities (\secref{sec:seed}).
The pre-defined hierarchy tree interacts with $\mathbf{v}_\text{map}$ through the hierarchy decomposition module \(\mathcal{D}\) such that the feature map is decomposed into the visual hierarchy (\secref{sec:decomp}).
Since the zero curvature of Euclidean space is not optimal for representing the hierarchical structure, we map the visual hierarchy to hyperbolic space and optimize the relation with a novel hierarchical contrastive loss (\secref{sec:hypbol}).
The visual hierarchy is then encoded back to the global visual representation $\mathbf{v}_\text{cls}$ through the hierarchy encoding module \(\mathcal{G}\) resulting in an enhanced global representation (\secref{sec:encod}).

\subsection{Probabilistic hierarchy tree}\label{sec:seed}
The main problem of the recent hierarchy-aware ViTs~\cite{quadtree,ding2023visual,ke2023learning} is that they define the hierarchical relations between the image tokens mainly through the self-attention scores.
Such a symmetric measurement is suboptimal for representing the asymmetric inclusive relation of parent-child nodes.
To handle the problem, we propose to define $L$ levels of hierarchy tree $\mathbf{T}$ by modeling each hierarchy node with a probability distribution.

Specifically, we first parameterize each leaf-level (at the initial level) node as a unique Gaussian distribution and subsequently define the higher-level node as a Mixture-of-Gaussians (MoG) of its corresponding child nodes.
Accordingly, the mean vector represents the cluster center of the visual semantic and the covariance captures the scale of each semantic cluster.

\vspace{-10pt}
\paragraph{Initial level.}
Let $\mathbf{C}^{1}=\{\mathbf{c}^{1}_{n}\}_{n=1}^{N_1}$ be a set of $N_1$ initial level nodes.
We parameterize each node \(\mathbf{c}^{1}_{n}\) as a normal distribution with a mean vector \(\mu^{1}_{n}\) and a diagonal covariance matrix \(\sigma^{1}_{n}\mathbf{I}\) in $\mathbb{R}^d$ as:
\begin{equation}
    p(\mathbf{c}^{1}_{n})\sim\mathcal{N}(\mu^{1}_{n},\sigma^{1}_{n}\mathbf{I}),
\end{equation}
where $\mu^{1}_{n}$ and $\sigma^{1}_{n}$ are randomly initialized.
We use reparameterization trick~\cite{reparam} for stable sampling, such that:
\begin{equation}\label{eq:reparam}
    \mathbf{c}^{1}_{n} = \mu^{1}_{n} + \epsilon*\sigma^{1}_{n}\in\mathbb{R}^d,
\end{equation}
where $\epsilon \sim \mathcal{N}(0, \mathbf{I})$.

\vspace{-10pt}
\paragraph{Subsequent level.}
We derive the remaining $(L-1)$ levels of hierarchy tree by conditioning each level on the preceding hierarchy level.
For each $l$-th level, we formulate a set of $N_l$ nodes $\mathbf{C}^{l}=\{\mathbf{c}^{l}_{k}\}_{k=1}^{N_l}$.
Concretely, the $k$-th node $\mathbf{c}^{l}_{k}$ at the $l$-th level is approximated by a MoG of its two corresponding child nodes, $\mathbf{c}^{l-1}_{2k-1}$ and $\mathbf{c}^{l-1}_{2k}$, as:
\begin{equation}\label{eq:mog}
    p(\mathbf{c}^{l}_{k})\sim\sum_{k{'}=2k-1}^{2k}\mathcal{N}(\mu^{l-1}_{k'},\sigma^{l-1}_{k'}\mathbf{I}).
\end{equation}
To stabilize the construction of hierarchy tree, we increase the sampling rate as the distribution expands, \ie, we sample $2^{l-1}$ embeddings from $p(\mathbf{c}^{l}_{k})$ using the reparameterization trick in \equref{eq:reparam} as:
\begin{equation}\label{eq:hier_node}
\begin{aligned}
\mathbf{c}^l_k = \{\hat{\mathbf{c}}^{l}_{k,1},...,\hat{\mathbf{c}}^{l}_{k,2^{l-1}}\} \overset{\text{iid}}{\sim} p(\mathbf{c}_k^l), \quad \mathbf{c}^l_k \in \mathbb{R}&^{2^{l-1}\times d}, \\
\mathbf{C}^{l} = \{\mathbf{c}^l_1,...,\mathbf{c}^l_{N_l}\} \in \mathbb{R}^{2^{l-1}N_l\times d}.
\end{aligned}
\end{equation}
By sequentially conditioning the higher-level nodes on the preceding lower-level nodes, we obtain the hierarchy tree $\mathbf{T}=\{\mathbf{C}^{l}\}^{L}_{l=1}$, where the lower-level nodes capture fine details with concentrated distribution and the higher-level nodes capture coarse instance-level representations with dispersed distribution.

\vspace{-10pt}
\paragraph{KL divergence loss.}
To prevent the variances from collapsing to zero, we employ KL regularization term between the distributions of $L$-th level nodes and the unit Gaussian prior \(\mathcal{N}(0,\mathbf{I})\) following~\cite{chun2021probabilistic}:
\begin{equation}
    \mathcal{L}_\text{KL} = \sum_{i=1}^{N_{L}} \mathrm{KL}(\mathcal{N}(\mu^{L}_{i}, \sigma^{L}_{i})\parallel\mathcal{N}(0, \mathbf{I})).
\end{equation}

\subsection{Visual hierarchy decomposition}
\label{sec:decomp}

Given the pre-defined hierarchy tree $\mathbf{T}$ and the visual feature map $\mathbf{v}_\text{map}$, we decompose $\mathbf{v}_\text{map}$ into $L$ levels of visual hierarchy through hierarchy decomposition module $\mathcal{D}$, as shown in \figref{fig:3a}.
We instantiate $\mathcal{D}$ as a stack of two transformer decoder layers.

To identify the visual hierarchy at the $l$-th level, the decomposition module $\mathcal{D}$ treats $\mathbf{C}^{l}$ as the query, and $\mathbf{v}_\text{map}$ as the key and value such that the semantically-relevant visual features are aggregated to the closest semantic cluster:
\begin{equation}\label{eq:semnodes}
    \mathbf{S}^{l} = \{\mathbf{s}^l_1,...,\mathbf{s}^l_{N_l}\} = \mathcal{D}(\mathbf{C}^{l}, \mathbf{v}_\text{map}) \in \mathbb{R}^{2^{l-1}N_l\times d}.
\end{equation}
Similar to the hierarchy nodes $\mathbf{c}_k^l$ in \equref{eq:hier_node}, the decomposed visual hierarchy nodes $\mathbf{s}^{l}_{k}$ are comprised of $2^{l-1}$ visual representations as $\mathbf{s}^l_k = \{\hat{\mathbf{s}}^{l}_{k,1},...,\hat{\mathbf{s}}^{l}_{k,2^{l-1}}\}$.
We average the set of $2^{l-1}$ representations for each visual hierarchy node $\mathbf{s}^{l}_{k}$ such that:
\begin{equation}
    \mathbf{s}^{l}_{k} = \frac{1}{2^{l-1}}\sum_{i=1}^{2^{l-1}} \hat{\mathbf{s}}^{l}_{k,i}, \quad {\mathbf{s}}^{l}_{k} \in \mathbb{R}^{d} 
\end{equation}
We perform the same decomposition procedure for $L$ levels, thereby obtaining $L$ levels of visual hierarchy in Euclidean space $\mathbf{T}_\mathbb{E}=\{\mathbf{S}^{l}\}^{L}_{l=1}$.

\subsection{Learning hierarchy in hyperbolic space}\label{sec:hypbol}
A natural characteristic of the hierarchical structure is that the number of nodes exponentially increases as the depth increases.
In practice, representing this property in Euclidean space leads to distortions in the semantic distances due to its flat geometry.
We propose to handle the problem by learning the hierarchical relations in hyperbolic space, where its exponentially expanding volume can efficiently represent the visual hierarchy.

We first map $\mathbf{T}_{\mathbb{E}}$ to the Lorentz hyperboloid to derive the visual hierarchy in hyperbolic space \(\mathbf{T}_{\mathbb{L}}\).
Following ~\cite{hypmeru}, we simplify the mapping computation by parameterizing only the space component of the Lorentz model as $\mathbf{s} = \left[{\mathbf{s}}^{l}_{k}, 0\right] \in \mathbb{R}^{d+1}$, where $\mathbf{s}$ belongs to the tangent space at the hyperboloid origin $\mathbf{O}$.
The visual hierarchy node in hyperbolic space $\mathbf{h}^{l}_{k} = [\mathbf{h}^{l}_{k,\text{space}}, {h}^{l}_{k,\text{time}}]$ is then obtained by transforming \(\mathbf{s}^{l}_{k}\) to $\mathbf{h}_{k,\text{space}}^{l}$ using the exponential map in \equref{eq:expm} as:
\begin{align}
    \label{eq:expm}
    \mathbf{h}_{k,\text{space}}^{l} =& \cosh({\sqrt{c}\lVert\mathbf{s}\rVert_{\mathbb{L}}})\mathbf{0}+\frac{\sinh(\sqrt{c}\lVert\mathbf{s}\rVert_{\mathbb{L}})}{\sqrt{c}\lVert\mathbf{s}\rVert_{\mathbb{L}}}\mathbf{s}_{k}^{l} \notag \\
    =& \frac{\sinh(\sqrt{c}\lVert\mathbf{s}^{l}_{k}\rVert)}{\sqrt{c}\lVert\mathbf{s}^{l}_{k}\rVert}\mathbf{s}^{l}_{k},
\end{align}
and computing the corresponding time component ${h}^{l}_{k,\text{time}}$ using \equref{eq:time} as ${h}^l_{k,\text{time}} = \sqrt{1/c+\lVert \mathbf{h}^l_{k,\text{space}}\rVert^{2}}$.
The visual hierarchy in hyperbolic space \( \mathbf{T}_{\mathbb{L}} \) is denoted as:
\begin{equation}
    \mathbf{T}_{\mathbb{L}} = \{\mathbf{H}^{l}\}_{l=1}^{L}, \quad \mathbf{H}^{l}=\{\mathbf{h}^{l}_{1},...,\mathbf{h}^{l}_{N_{l}}\}\in\mathbb{R}^{N_{l}\times d_{\mathbb{L}}},
\end{equation}
where \(d_{\mathbb{L}}\) is the embedding dimension in Lorentz manifold.

\vspace{-10pt}
\paragraph{Hierarchical contrastive loss.}

To guarantee the hierarchy nodes to reflect the hierarchical relations, we optimize the distances of \(\mathbf{T}_{\mathbb{L}}\) with a novel hierarchical contrastive loss $\mathcal{L}_{\mathbb{L}\text{-cont}}$.
Specifically, we formulate the similarity between the nodes in consideration of the length of the connected branch, \ie, \textit{geodesic} distance, as shown in \figref{fig:2b}.

Consider a node $\mathbf{h}_{2k}^l$ in \(\mathbf{T}_{\mathbb{L}}\), where $\mathbf{h}_{k}^{l+1}$ is its parent node.
We encourage $\mathbf{h}_{2k}^l$ to be similar with $\mathbf{h}_{k}^{l+1}$ since the pair of parent-child nodes lie on the same branch.
Meanwhile, we encourage the remaining nodes in the same level, \ie, $\{\mathbf{h}_i^l|\mathbf{h}_i^l\in\mathbf{H}^l, i\neq 2k\}$, to be dissimilar since they all lie on separate branches.
The loss also penalizes the close sibling node $\mathbf{h}^{l}_{2k-1}$ as their geodesic traverses the parent node. 
Formally, the hierarchical contrastive loss incorporates the geometric interpretation of hierarchical structure into the contrastive loss~\cite{infonce} as:
\begin{equation}
    \mathcal{L}_{\mathbb{L}\text{-cont}}=-\log\underset{\mathbf{T}_{\mathbb{L}}}{\mathbb{E}}\left[\sum_{i=0}^1\frac{\exp(D_{\mathbb{L}}(\mathbf{h}_{k}^{l+1},\mathbf{h}_{2k-i}^{l}))}{\sum_{j\neq 2k-i}\exp(D_{\mathbb{L}}(\mathbf{h}_{2k-i}^{l},\mathbf{h}_{j}^{l}))}\right],
\end{equation}
where \(D_{\mathbb{L}}\) is the Lorentzian distance defined in \equref{eq:dist}.

By regarding every pair of parent-child nodes as positive and every pair of nodes at the same level as negative, we are able to represent the logical structure of visual hierarchy without the part-level annotation.
In addition, the hierarchical relation optimized through $\mathbf{T}_\mathbb{L}$ in hyperbolic space is well-preserved in $\mathbf{T}_\mathbb{E}$ in Euclidean space thanks to the mapping computation in \equref{eq:expm}.

\subsection{Visual hierarchy encoding}\label{sec:encod}

We propose to enhance the structured understanding of the global visual representation by encoding the identified visual hierarchy into $\textbf{v}_\text{cls}$.
While we optimize the hierarchical relations in hyperbolic space due to its ability of handling \textit{relative} distances, its complex computations and expansion property make it less suitable for tasks that require \textit{absolute} criteria.
Thus, we exploit the visual hierarchy in Euclidean space $\mathbf{T}_\mathbb{E}$ for hierarchy encoding. 

Specifically, we progressively encode the hierarchy nodes $\mathbf{S}^{l}$ into $\textbf{v}_\text{cls}$, from the initial level to the $L$-th level, via the hierarchy encoding module $\mathcal{G}$ such that $\textbf{v}_\text{cls}$ is updated as a hierarchy-aware global representation $\textbf{v}_\text{cls}^L$.
As shown in \figref{fig:3b}, the module $\mathcal{G}$ is instantiated as a stack of $L$ transformer decoder layers.

To encode the $l$-th level hierarchy, the $l$-th layer of $\mathcal{G}^l$ takes $\textbf{v}_\text{cls}^{l-1}$ as the query, and $\mathbf{S}^{l}$ as the key and value:
\begin{equation}
    \mathbf{v}^{l}_\text{cls} = \mathcal{G}^{l}(\mathbf{v}^{l-1}_{\text{cls}}, \mathbf{S}^{l}) \in \mathbb{R}^d, \quad 0< l \leq L, 
\end{equation}
where we set \(\mathbf{v}^{0}_{\text{cls}}\) as \(\mathbf{v}_{\text{cls}}\).
The enhanced global visual representation is then derived as:
\begin{equation}
    \hat{\mathbf{v}}_\text{cls} = \mathbf{v}_{\text{cls}} + \mathbf{v}^{L}_{\text{cls}}.
\end{equation}
For the final prediction, we feed $\hat{\mathbf{v}}_\text{cls}$ into the pre-trained classifier of $\mathcal{F}$. 

\vspace{-10pt}
\paragraph{Overall objectives.}
The overall objective is defined as:
\begin{equation}
    \mathcal{L}_\text{total} = \mathcal{L}_\text{CE} + \alpha\mathcal{L}_{\mathbb{L}\text{-cont}}+\beta\mathcal{L}_\text{KL},
\end{equation}
where \(\alpha\) and \(\beta\) are the balancing parameters.

%% file: sec/5_experiment.tex
\section{Experiments}
\label{sec:experiment}
In this section, we conduct extensive experiments to show the effectiveness of our proposed Hi-Mapper on image classification (\secref{sec:cls}), object detection and instance segmentation (\secref{sec:object}), and semantic segmentation (\secref{sec:seg}).
We apply our Hi-Mapper on both CNN-based~\cite{resnet, efficientnet} and ViT-based~\cite{deit, swin} backbone networks and compare the performance with the concurrent hierarchy-aware baselines~\cite{crossvit,ding2023visual}.
Lastly, we provide ablation studies (\secref{sec:abbl}) to demonstrate the effectiveness of our contributions.

\input{tab/table_main}
\subsection{Image classification}\label{sec:cls}
\paragraph{Settings.}
We apply our Hi-Mapper on the state-of-the-art backbone networks~\cite{resnet,efficientnet,deit,swin} and benchmark on ImageNet-1k~\cite{deng2009imagenet} dataset.
Following~\cite{davit,zhang2021multi,lin2017focal,swin,cvt}, we use the identical combinations for data augmentation and regularization strategies~\cite{deit} after excluding repeated augmentation~\cite{hoffer2020augment}.
We train our model with batch size 1024 for 50 epochs using AdamW~\cite{loshchilov2017decoupled} with weight decay 1e-4.
The initial learning rate is set to 1e-4 and a cosine learning rate scheduler is applied following ~\cite{deit}.
\vspace{-10pt}
\paragraph{Results.}
\tabref{tab:main} presents classification performance.
We report the original performance without the fine-tuning schemes for the plain backbone networks since we observed degradation in performance after fine-tuning the models.
Our proposed method consistently achieves better performances than the baseline models with only a slight increase in parameters.
Specifically, Hi-Mapper surpasses ResNet-50~\cite{resnet} and EfficientNet-B4~\cite{efficientnet} by margins of 2.0\% and 1.2\%, respectively. 
Additionally, it achieves improvement on DeiT's performance by 2.6\%/2.8\%/1.6\%, and Swin~\cite{swin} by 2.2\%/1.1\% across various model sizes.
The experiments on image classification demonstrate not only the importance of understanding the structured organization of visual scenes, but also the scalability of our method.

\input{tab/tab_mascrcnn}

\subsection{Object detection and Instance segmentation}\label{sec:object}
\paragraph{Settings.}
We experiment on the COCO~\cite{lin2014microsoft} dataset. 
We use the Hi-Mapper backbones derived from~\secref{sec:cls}.
Then, we deploy our pre-trained backbone into Mask R-CNN~\cite{he2017mask}.
All models are trained on COCO 2017, including 118k train images and 5k validation images.
We follow the standard learning protocols~\cite{he2017mask}, 1\(\times\) schedule with 12 epochs.
Note that we reproduce DeiT-T with Mask R-CNN based on~\cite{li2022exploring}.

\input{tab/tab_seg}
\vspace{-10pt}
\paragraph{Results.}
In Table~\ref{tab:maskrcnn}, we present a comparison of our results with baseline models, such as DeiT and Swin, on the COCO dataset, which demands a higher capacity for fine representation recognition.
Our Hi-Mapper consistently boosts all baseline models capability with only a small increase in parameters.
Notably, Hi-Mapper significantly improves DeiT-T and -S by margins of 6.8 and 8.3 on the object detection task.
Meanwhile, in the instance segmentation, it also improves DeiT-T and -S by margins of 8.6 and 7.2, respectively.
This result shows that the visual hierarchy facilitates complex visual scene recognition.
\subsection{Semantic segmentation}\label{sec:seg}
\paragraph{Settings.}
We further experiment on the ADE20K~\cite{zhou2017scene} dataset for semantic segmentation.
ADE20K contains 20k training images, 20K validation images, and 3K test images, covering a total 150 classes.
Following common practice~\cite{xiao2018unified}, we report the mIoU on the validation set. 
\input{tab/table_EH}
\vspace{-10pt}
\paragraph{Results.}
We present the performance comparisons on ADE20K in~\tabref{tab:seg}.
The results show that our Hi-Mapper achieves comparable or better performance than the baseline models, including DeiT-T, -S, and Swin-T, requiring a smaller increase in the number of parameters and GFLOPs.
Specifically, Hi-Mapper on DeiT-T, -S, and Swin-T achieves a performance improvement of 2.0\%, 3.3\%, and 2.3\%. 
Additionally, as model sizes increase, Hi-Mapper fully capitalizes fine-grained representations for semantic segmentation with a slight increase in computation.

\subsection{Visualization}
\begin{figure*}[!t]
\label{fig:visual}
    \centering
    \includegraphics[width=0.95\linewidth]{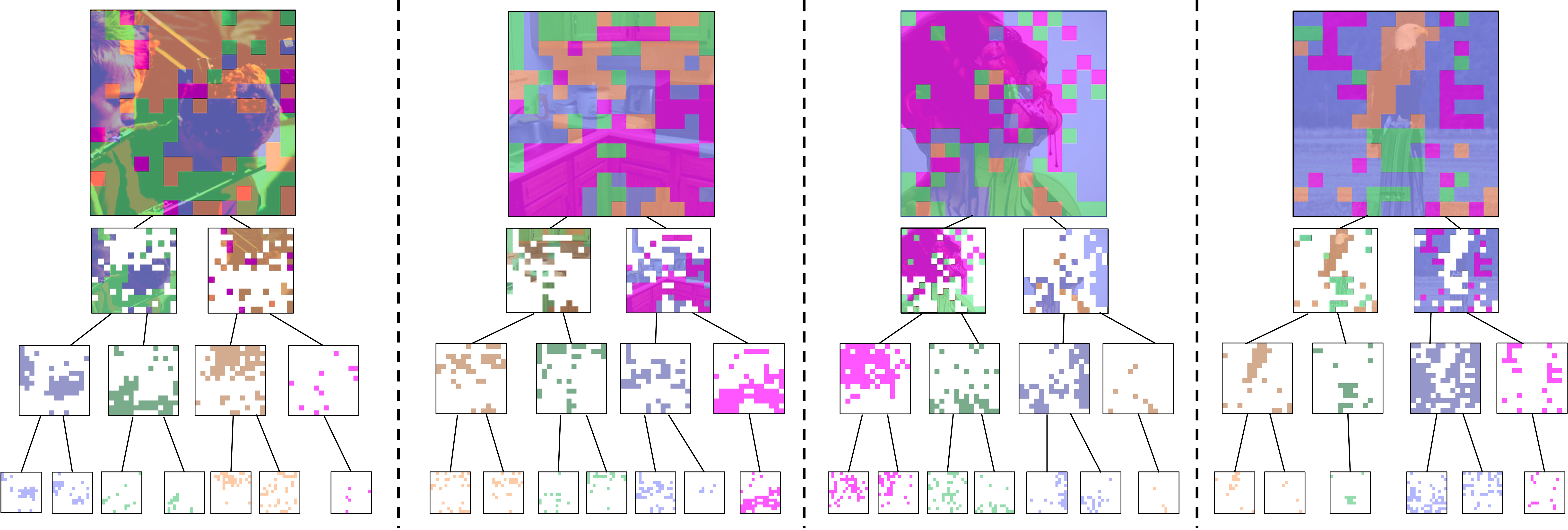}
    \vspace{-5pt}
    \caption{Visualization of visual hierarchy decomposed by Hi-Mapper(Deit-S) trained on ImageNet-1K with classification objective. 
    Each color represents different subtrees.
    We ignore the nodes of the small region and display only the main subtrees.}
    \label{fig:vis}
    \vspace{-10pt}
\end{figure*}

As shown in~\figref{fig:vis}, we demonstrate our visual hierarchy on images from ImageNet-1K~\cite{deng2009imagenet}.
This confirms that our approach can successfully uncover the inherent hierarchy among visual components without the need for hierarchy or part-level annotations.

\section{Ablation studies and discussion}\label{sec:abbl}
To further analyze and validate the components of our method, we conduct ablation studies on image classification.
\vspace{-5pt}
\input{tab/table_seed}
\vspace{-15pt}
\paragraph{Effectiveness of hyperbolic manifolds.}
We first investigate the effectiveness of hyperbolic manifolds in our approach. 
As shown in~\tabref{tab:eh}, we report the impact according to image classification.
In Euclidean space, the distance function between two vectors is the cosine similarity function.
The results demonstrate that applying hierarchical contrastive loss in Euclidean space degrades performance. 
It indicates that hyperbolic space is more suitable for stabilizing hierarchical structures. 
Additionally, the application of a KL loss term shows further benefits derived from the semantic seed distribution.
\vspace{-10pt}
\paragraph{Impact of probabilistic modeling.}
In~\tabref{tab:seed}, we report the performance comparisons between the probabilistic hierarchy tree and the deterministic hierarchy tree.
For constructing a hierarchy tree, probabilistic modeling defines every node via MoG of its child node distributions, while the deterministic approach determines each node the mean of its child nodes.
The probabilistic hierarchy tree achieves significant improvement in performance compared to the deterministic approach.
This result shows that probabilistic modeling is more effective in representing hierarchical structure than deterministic modeling, leading to improvement in recognition.
\vspace{-10pt}
\paragraph{Hierarchy width and depth.}
As shown in~\figref{fig:abla}, we analyze the effect of the width \(N\) and depth \(L\) of the hierarchy tree on ImageNet-1K~\cite{deng2009imagenet} with Hi-Mapper(DeiT-S).
These factors control the granularity of visual elements to be decomposed.
While a small number of \(N\) degrades the fine-grained recognition capacity, an excess \(N\) hinders the optimization.
Meanwhile, a large \(L\) may provide diverse granularity, however, it leads to entangled object-level representations.
In all the cases, we report the best performance of 82.6\% at \(N=32, L=4\).
\input{figure/graph}

%% file: tab/table_main.tex
\definecolor{LinkWater}{rgb}{0.85,0.882,0.949}
\begin{table}
\centering
\caption{Performance comparisons for image classification on ImageNet-1K~\cite{deng2009imagenet} dataset.} 

\label{tab:main}
{\small
\resizebox{0.95\linewidth}{!}{
\begin{tblr}{
  width = \linewidth,
  colspec = {Q[80, leftsep=1pt, rightsep=1pt]Q[280, leftsep=3pt, rightsep=1pt]Q[120, leftsep=1pt, rightsep=1pt]Q[120, leftsep=1pt, rightsep=1pt]Q[120, leftsep=1pt, rightsep=1pt]},
  row{2} = {c},
  row{5} = {LinkWater},
  row{6} = {LinkWater},
  row{11} = {LinkWater},
  row{17} = {LinkWater},
  row{18} = {LinkWater},
  row{21} = {LinkWater},
  row{22} = {LinkWater},
  column{1} = {c},
  cell{1}{1} = {r=2}{c},
  cell{1}{2} = {r=2}{},
  cell{1}{3} = {c},
  cell{1}{4} = {c},
  cell{1}{5} = {c},
  cell{3}{1} = {r=4}{c},
  cell{3}{3} = {c},
  cell{3}{4} = {c},
  cell{3}{5} = {c},
  cell{4}{3} = {c},
  cell{4}{4} = {c},
  cell{4}{5} = {c},
  cell{5}{3} = {c},
  cell{5}{4} = {c},
  cell{5}{5} = {c},
  cell{6}{3} = {c},
  cell{6}{4} = {c},
  cell{6}{5} = {c},
  cell{7}{1} = {r=16}{},
  cell{7}{3} = {c},
  cell{7}{4} = {c},
  cell{7}{5} = {c},
  cell{8}{3} = {c},
  cell{8}{4} = {c},
  cell{8}{5} = {c},
  cell{9}{3} = {c},
  cell{9}{4} = {c},
  cell{9}{5} = {c},
  cell{10}{3} = {c},
  cell{10}{4} = {c},
  cell{10}{5} = {c},
  cell{11}{3} = {c},
  cell{11}{4} = {c},
  cell{11}{5} = {c},
  cell{12}{3} = {c},
  cell{12}{4} = {c},
  cell{12}{5} = {c},
  cell{13}{3} = {c},
  cell{13}{4} = {c},
  cell{13}{5} = {c},
  cell{14}{3} = {c},
  cell{14}{4} = {c},
  cell{14}{5} = {c},
  cell{15}{3} = {c},
  cell{15}{4} = {c},
  cell{15}{5} = {c},
  cell{16}{3} = {c},
  cell{16}{4} = {c},
  cell{16}{5} = {c},
  cell{17}{3} = {c},
  cell{17}{4} = {c},
  cell{17}{5} = {c},
  cell{18}{3} = {c},
  cell{18}{4} = {c},
  cell{18}{5} = {c},
  cell{19}{3} = {c},
  cell{19}{4} = {c},
  cell{19}{5} = {c},
  cell{20}{3} = {c},
  cell{20}{4} = {c},
  cell{20}{5} = {c},
  cell{21}{3} = {c},
  cell{21}{4} = {c},
  cell{21}{5} = {c},
  cell{22}{3} = {c},
  cell{22}{4} = {c},
  cell{22}{5} = {c},
  cell{23}{3} = {c},
  cell{23}{4} = {c},
  cell{23}{5} = {c},
  vline{2,3} = {-}{},
  hline{1,3,7,23} = {-}{},
  hline{12,19} = {2-5}{},
}
Type        & Model                                                                                                                                               & Params & FLOPs & Top-1                  \\
            &                                                                                                                                                     & (M)    & (G)   & (\%)                   \\
{CNN}        
            & ResNet-50~\cite{resnet}                                                                                                                                           & 25.6   & 3.8   & 76.2                   \\
            & EfficientNet-B4*~\cite{efficientnet}                                                                                                                                    & 19.3   & 4.2   & 82.9                   \\
            & Hi-Mapper(RN50)                            & 26.9   & 4.0   & \textbf{78.2}                   \\
            & Hi-Mapper(ENB4*)                    & 20.5   & 4.4   & \textbf{84.1}                   \\
{ViT} & DeiT-T~\cite{deit}                                                                                                                                              & 5.7    & 1.3   & 72.2                   \\
            & CrossViT-T~\cite{crossvit}                                                                                                                                          & 6.9    & 1.6   & 73.4                   \\
            & PVT-T~\cite{wang2022pvt}                                                                                                                                               & 13.2   & 1.9   & 75.1                   \\
            & DependecnyViT-T~\cite{ding2023visual}                                                                                                                                     & 6.2    & 1.3   & \textbf{75.4}          \\
            & Hi-Mapper(DeiT-T)                            & 6.6    & 1.5   & 74.8                   \\
            & DeiT-S~\cite{deit}                                                                                                                                              & 22.2   & 4.5   & 79.8                   \\
            & CrossViT-S~\cite{crossvit}                                                                                                                                             & 26.7   & 5.6   & 81                     \\
            & PVT-S~\cite{wang2022pvt}                                                                                                                                               & 24.5   & 3.8   & 79.8                   \\
            & Swin-T~\cite{swin}                                                                                                                                              & 28.3   & 4.6   & 81.2                   \\
            & DependencyViT-S~\cite{ding2023visual}                                                                                                                                     & 24.0     & 5.0     & 82.1                   \\
           
            & Hi-Mapper(DeiT-S)             & 23.3   & 4.8   & \textbf{82.6} \\
             & Hi-Mapper(Swin-T)     & 29.5   & 4.9   & \textbf{\textbf{83.4}} \\
             & DeiT-B~\cite{deit}    & 85.6   & 17.6   & 81.8                   \\
             & Swin-S~\cite{swin}    & 50.1   & 8.7   & 83.0                   \\
             & Hi-Mapper(DeiT-B)     & 87.2   & 18.1   & \textbf{83.4} \\
             & Hi-Mapper(Swin-S)     & 51.8   & 9.3   & {\textbf{84.1}}
\end{tblr}}}
\vspace{-10pt}
\end{table}

%% file: tab/tab_mascrcnn.tex
\begin{table*}[t]
\centering
{\small
\caption{Performance comparisons for object detection and instance segmentation on COCO~\cite{lin2014microsoft} dataset.(*: Results
are reproduced based on~\cite{zhao2020exploring}.}
\vspace{-5pt}
\label{tab:maskrcnn}
\resizebox{0.95\linewidth}{!}{
\begin{tblr}{
  width = \linewidth,
  colspec = {Q[200]Q[69]Q[60]Q[110]Q[110]Q[110]Q[110]Q[110]Q[110]},
  row{2} = {c},
  row{5} = {LinkWater},
  row{9} = {LinkWater},
  row{10} = {LinkWater},
  column{2} = {c},
  column{3} = {c},
  cell{1}{1} = {r=2}{},
  cell{1}{2} = {r=2}{},
  cell{1}{3} = {r=2}{},
  cell{1}{4} = {c=6}{c},
  cell{3}{4} = {c},
  cell{3}{5} = {c},
  cell{3}{6} = {c},
  cell{3}{7} = {c},
  cell{3}{8} = {c},
  cell{3}{9} = {c},
  cell{4}{4} = {c},
  cell{4}{5} = {c},
  cell{4}{6} = {c},
  cell{4}{7} = {c},
  cell{4}{8} = {c},
  cell{4}{9} = {c},
  cell{5}{4} = {c},
  cell{5}{5} = {c},
  cell{5}{6} = {c},
  cell{5}{7} = {c},
  cell{5}{8} = {c},
  cell{5}{9} = {c},
  cell{6}{4} = {c},
  cell{6}{5} = {c},
  cell{6}{6} = {c},
  cell{6}{7} = {c},
  cell{6}{8} = {c},
  cell{6}{9} = {c},
  cell{7}{4} = {c},
  cell{7}{5} = {c},
  cell{7}{6} = {c},
  cell{7}{7} = {c},
  cell{7}{8} = {c},
  cell{7}{9} = {c},
  cell{8}{4} = {c},
  cell{8}{5} = {c},
  cell{8}{6} = {c},
  cell{8}{7} = {c},
  cell{8}{8} = {c},
  cell{8}{9} = {c},
  cell{9}{4} = {c},
  cell{9}{5} = {c},
  cell{9}{6} = {c},
  cell{9}{7} = {c},
  cell{9}{8} = {c},
  cell{9}{9} = {c},
  cell{10}{4} = {c},
  cell{10}{5} = {c},
  cell{10}{6} = {c},
  cell{10}{7} = {c},
  cell{10}{8} = {c},
  cell{10}{9} = {c},
  vline{2,4} = {1-2,3-10}{},
  hline{1,11} = {-}{0.15em},
  hline{3,6} = {-}{0.05em},
  hline{2} = {3-9}{0.05em},
}
Backbone          & {Params\\(M)} & {Flops\\(G)} & Mask R-CNN 1x &               &               &               &               &               \\
                  &               &              & $\text{AP}^\text{box}$        & $\text{AP}^\text{box}_{50}$     & $\text{AP}^\text{box}_{75}$     & $\text{AP}^\text{mask}$        & $\text{AP}^\text{mask}_{50}$     & $\text{AP}^\text{mask}_{75}$     \\
PVT-T~\cite{wang2022pvt}            & 32.9          & 195       & 36.7          & 59.2          & 39.3          & 35.1          & 56.7          & 37.3          \\
DeiT-T*~\cite{deit}       & 27.3          & 244       & 30.3          & 46.2          & 32.1          & 27.1          & 44.4          & 28.3          \\
Hi-Mapper(Deit-T) & 29.5          & 246       & 37.1 & 61.7 & 41.0 & 35.7 & 59.1 & 38.1 \\
PVT-S~\cite{wang2022pvt}            & 44.1          & 245       & 40.4          & 62.9          & 43.8          & 37.8          & 60.1          & 40.3          \\
DeiT-S*~\cite{deit}           & 44.9          & 276      & 36.3          & 54.1          & 39.0          & 31.7          & 51.3          & 33.2          \\
Swin-T~\cite{swin}            & 47.8          & 267      & 43.5          & 66.4          & 47.3          & 39.6          & 63.1          & 42.3          \\
Hi-Mapper(Deit-S) & 47.0          & 279      & 42.6 & 65.8 & 46.3 & 38.9 & 62.7 & 41.8 \\
Hi-Mapper(Swin-T) & 50.4          & 270      & 44.0          & 67.1          & 47.9          & 39.9          & 63.6          & 42.8          
\end{tblr}}}
\vspace{-10pt}
\end{table*}

%% file: tab/tab_seg.tex
\begin{table}[t]
\centering
\setlength{\extrarowheight}{0pt}
\addtolength{\extrarowheight}{\aboverulesep}
\addtolength{\extrarowheight}{\belowrulesep}
\setlength{\aboverulesep}{0pt}
\setlength{\belowrulesep}{0pt}
\caption{Performance comparisons for semantic segmentation on ADE20k~\cite{zhou2017scene} dataset. 
We conduct the single-scale evaluation.
FLOPs are measured with 512 \(\times\) 2048 input resolution.}
\vspace{-5pt}
\label{tab:seg}
\resizebox{0.99\linewidth}{!}{
\footnotesize{
\begin{tblr}{
  width = \linewidth,
  colspec = {Q[440]Q[320]Q[100]Q[100]Q[100]},
  row{2} = {c},
  row{6} = {LinkWater},
  row{11} = {LinkWater},
  row{12} = {LinkWater},
  cell{1}{1} = {r=2}{},
  cell{1}{2} = {r=2}{},
  cell{1}{3} = {c},
  cell{1}{4} = {c},
  cell{1}{5} = {r=2}{},
  cell{3}{3} = {c},
  cell{3}{4} = {c},
  cell{3}{5} = {c},
  cell{4}{3} = {c},
  cell{4}{4} = {c},
  cell{4}{5} = {c},
  cell{5}{3} = {c},
  cell{5}{4} = {c},
  cell{5}{5} = {c},
  cell{6}{3} = {c},
  cell{6}{4} = {c},
  cell{6}{5} = {c},
  cell{7}{3} = {c},
  cell{7}{4} = {c},
  cell{7}{5} = {c},
  cell{8}{3} = {c},
  cell{8}{4} = {c},
  cell{8}{5} = {c},
  cell{9}{3} = {c},
  cell{9}{4} = {c},
  cell{9}{5} = {c},
  cell{10}{3} = {c},
  cell{10}{4} = {c},
  cell{10}{5} = {c},
  cell{11}{3} = {c},
  cell{11}{4} = {c},
  cell{11}{5} = {c},
  cell{12}{3} = {c},
  cell{12}{4} = {c},
  cell{12}{5} = {c},
  vline{2,3} = {-}{},
  hline{1,3,7,13} = {-}{},
}
Backbone          & Method       & Params & FLOPs & mIoU \\
                  &              & (M)    & (G)   &      \\
PVT-T~\cite{wang2022pvt}             & SemanticFPN~\cite{kirillov2019panoptic} & 17.0          & 132       & 35.7          \\
DeiT-T~\cite{deit}            & UperNet~\cite{xiao2018unified}          & 10.7     & 142       & 37.8          \\
DependencyViT-T~\cite{ding2023visual}   & UperNet~\cite{xiao2018unified}          & 11.1        & 145       & 40.3 \\
Hi-Mapper(Deit-T) & UperNet~\cite{xiao2018unified}          & 11.6      & 144      & 39.8          \\
PVT-S~\cite{wang2022pvt}             & SemanticFPN~\cite{kirillov2019panoptic} & 28.2          & 712       & 39.8          \\
DeiT-S~\cite{deit}            & UperNet~\cite{xiao2018unified}          & 41.3      & 566       & 43.0          \\
Swin-T~\cite{swin}            & UperNet~\cite{xiao2018unified}          & 60.0        & 945        & 44.5          \\
DependencyVIT-S~\cite{ding2023visual}   & UperNet~\cite{xiao2018unified}          & 43.1        & 574       & 45.7          \\
Hi-Mapper(Deit-S) & UperNet~\cite{xiao2018unified}          & 42.5       & 570       & 46.3 \\
Hi-Mapper(Swin-T) & UperNet~\cite{xiao2018unified}          & 62.1      & 949       & 46.8
\end{tblr}}}
\vspace{-10pt}
\end{table}

%% file: tab/table_EH.tex
\begin{table}
\centering
\caption{Performance comparison for classification on ImageNet-1K~\cite{deng2009imagenet} according to embedding spaces and combinations of learning objectives.}
\vspace{-5pt}
\label{tab:eh}
\small{

\begin{tblr}{
  width = 0.90\linewidth,
  colspec = {Q[350]Q[259]Q[259]Q[259]},
  cells = {c},
  cell{2}{1} = {r=4}{},
  cell{6}{1} = {r=4}{},
  vline{2,4} = {-}{},
  hline{1-2,6,10} = {-}{},
  cell{9}{2,3,4} = {bg=LinkWater}
}
Manifold      & $\mathcal{L}_{\mathbb{L}\text{-cont}}$ & \(\mathcal{L}_\text{KL}\) & Top-1(\%)         \\
Euclidean  &  \xmark          &     \xmark              & 79.7          \\
           &  \xmark                   &       \cmark            & 79.6          \\
           &    \cmark                 &     \xmark              & 79.2          \\
           &    \cmark                 &       \cmark            & 79.3 \\
Hyperbolic &   \xmark                  & \xmark                  & 79.7          \\
           &   \xmark                  &       \cmark            & 79.5          \\
           &      \cmark               &      \xmark             & 82.0          \\
           &    \cmark                 &     \cmark              & 82.6
\end{tblr}}\vspace{-10pt}
\end{table}

%% file: tab/table_seed.tex
\begin{table}
\centering
\caption{Performance comparison for classification on ImageNet-1K~\cite{deng2009imagenet} with respect to the relation modeling in hierarchy tree.}\vspace{-5pt}
\label{tab:seed}
\resizebox{0.90\linewidth}{!}{
\begin{tblr}{
  width = \linewidth,
  colspec = {Q[450]Q[300]Q[300]},
  row{1} = {c},
  cell{2}{2} = {c},
  cell{2}{3} = {c},
  cell{2}{3} ={LinkWater},
  hline{1,3}={-}{1pt},
  hline{2}={-}{},
}
Model                         & Deterministic & Probabilistic \\
Hi-Mapper(DeiT-S) & 81.5\%           & 82.6\%        
\end{tblr}}\vspace{-10pt}
\end{table}

%% file: figure/graph.tex
\begin{figure}[!t]
\centering
  \begin{subfigure}{0.49\linewidth}
      \centering
      \includegraphics[width=1\linewidth]{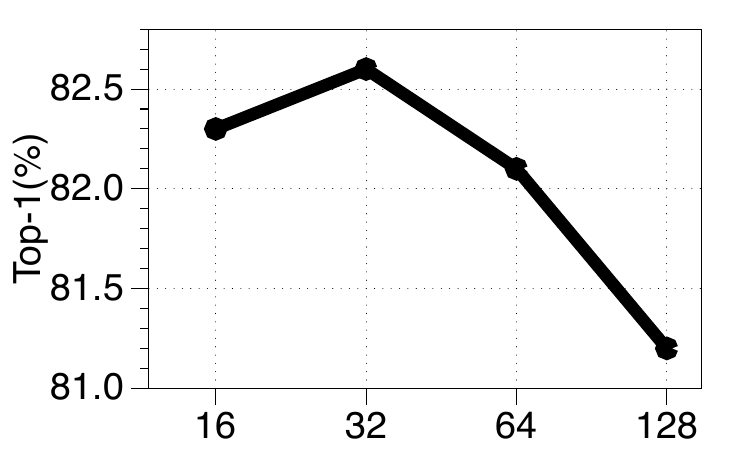}
      \caption{\(N\in\{16,32,64,128\}\)}
  \end{subfigure}
  \begin{subfigure}{0.49\linewidth}
      \centering
      \includegraphics[width=1\linewidth]{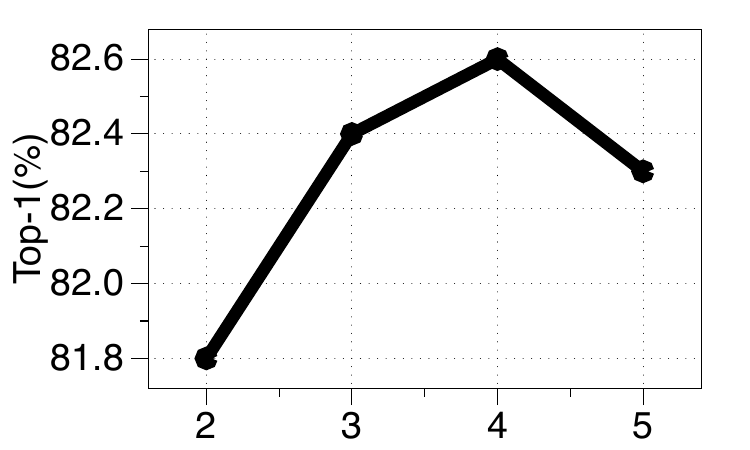}
      \caption{\(L\in\{2,3,4,5\}\)}
  \end{subfigure}\vspace{-5pt}
    \caption{Hyper-parameter analysis on ImageNet-1K.}
\vspace{-10pt}
\label{fig:abla}
\end{figure}

%% file: sec/6_conclusion.tex
\section{Conclusion}
In this paper, we have presented a novel Visual Hierarchy Mapper (Hi-Mapper) that investigates the hierarchical organization of visual scenes.
We have achieved the goal by newly defining tree-like structure with probability distribution and learning the hierarchical relations in hyperbolic space.
We have incorporated the hierarchical interpretation into the contrastive loss and efficiently identified the visual hierarchy in a data-efficient manner.
Through an effective hierarchy decomposition and encoding procedures, the identified hierarchy has been successfully deployed to the global visual representation, enhancing the structured understanding of an entire scene. 
Hi-Mapper has consistently improved the performance of the existing DNNs when integrated with them, and also has demonstrated the effectiveness on various dense predictions.
\vspace{-15pt}
\paragraph{Acknowledgement.}
This research was supported by the Yonsei Signature Research Cluster Program of 2022 (2022-22-0002).

%% file: supplementary.tex
\appendix
In this document, we include supplementary materials for ``Improving Visual Recognition with Hyperbolical Visual Hierarchy Mapping".
We first provide more concrete implementation details (\secref{supp:arc}), a theoretical baseline (\secref{supp:base}), and additional experimental results (\secref{supp:addres}).
Finally, we visualize more visual hierarchy trees from the selected images to provide solid evidence of the proposed method (\secref{supp:addvis}). 
\vspace{-10pt}
\section{Network Architecture}\label{supp:arc}
\subsection{Classification}
To demonstrate the scalability of our proposed method, we deploy our method on ResNet50~\cite{resnet}, EfficientNet~\cite{efficientnet}, DeiT~\cite{deit}, and Swin~\cite{swin}.
All encoders are pre-trained on ImageNet-1K~\cite{deng2009imagenet} dataset.
Then, the hierarchy decomposition module \(\mathcal{D}\) is composed of two transformer layer, and then the Hierarchy encoding module \(\mathcal{G}\) is composed of four transformer layers, in which each transformer layer has four attention heads with 128 embedding dimensions.
For hyperbolic embedding, we initialize curvature parameter \(c=1\).
For dense prediction tasks, including semantic segmentation and object detection, we feed \(\mathbf{v}_{\text{map}}\) into the hierarchy encoding stage instead of \(\mathbf{v}_{\text{cls}}\).
We obtain $\mathbf{v}_{\text{cls}}$ by applying global average pooling on $\mathbf{v}_{\text{map}}$ for ResNet~\cite{resnet} and Swin~\cite{swin}, and utilize the \texttt{[CLS]} embedding for DeiT~\cite{deit}.
\vspace{-10pt}
\subsection{Dense prediction.}
\begin{figure*}[h]
    \centering
    \includegraphics[width=0.95\textwidth]{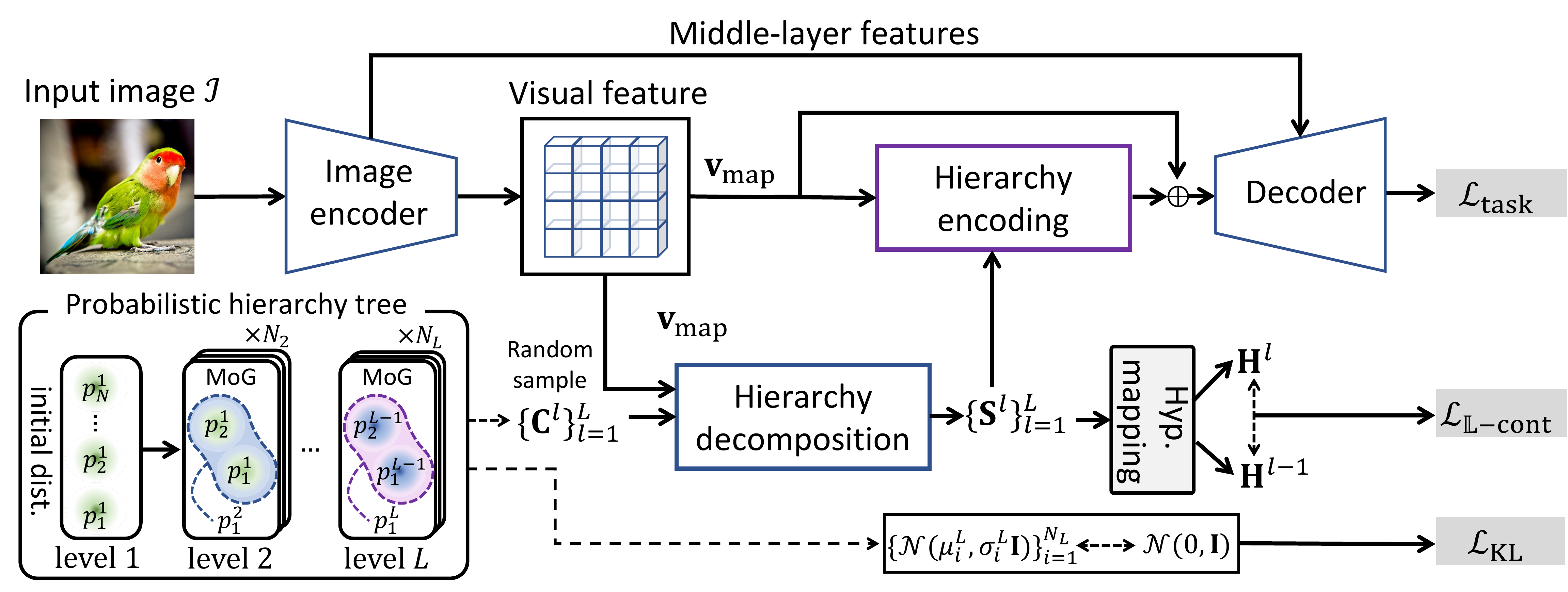}
    \caption{Illustration for overall procedure of Hi-Mapper for dense prediction tasks. 
     }
\vspace{-10pt}
\label{fig:dense}
\end{figure*}

In the main paper, we report utilizing UperNet~\cite{xiao2018unified} and Mask-RCNN~\cite{he2017mask} for semantic segmentation and object detection \& instance segmentation tasks. 
For each task's decoder, we incorporated the penultimate feature map \(\hat{\mathbf{v}}_\text{map}\), computed by our Hi-Mapper, along with the vanilla output from the intermediate layers of encoder \(\mathcal{F}\), as illustrated in~\figref{fig:dense}.
\section{Theoretical Baseline}\label{supp:base}
\subsection{Mixture of Gaussians}
For hierarchy tree \(\mathbf{T}\), each node distribution at level \(l+1\) is represented as a Mixture of Gaussians (MoG) of its child node distributions, which are independent. 
Hence, we can derive the semantic seed distribution at level \(l+1\) through a simple computation.
\begin{equation}\nonumber
    f_{k}^{l+1}(z)
    =\frac{1}{2}\sum_{i=0}^{1}f_{2k-i}^{l}(z),
\end{equation}
where \(f_{k}^{l+1}\) is the PDF for \(\mathbf{c}_{k}^{l+1}\).
Then, the mean of the MoG is formulated follow as:
\input{supple/fvsft}
\begin{align}
    \mu_{k}^{l+1}\nonumber
    &=\int zf_{k}^{l+1}(z)dz\\\nonumber
    &=\frac{1}{2}\sum_{i=0}^{1}\int zf_{2k-i}^{l}(z)dz\\\nonumber
    &=\frac{1}{2}\sum_{i=0}^{1}\mu_{2k-i}^{l}.\\\nonumber
\end{align}
The standard deviation $(\sigma_{k}^{l+1})^2$ is derived as follow:
\begin{align}
    (\sigma_{k}^{l+1})^2\nonumber
    &=\int z^2f_{k}^{l+1}(z)dz - (\mu_{k}^{l+1})^2\\\nonumber
    &=\frac{1}{2}\sum_{i=0}^{1}\int z^2f_{2k-i}^{l}(z)dz - (\mu_{k}^{l+1})^2\\\nonumber
    &=\frac{1}{2}\sum_{i=0}^{1}((\mu_{2k-i}^{l})^2+(\sigma_{2k-i}^{l})^2) - (\mu_{k}^{l+1})^2.\\\nonumber
\end{align}
\vspace{-15pt}
\section{Additional Results}\label{supp:addres}
\subsection{Fine-tuning vs. full-training.}
We also investigate the effectiveness of our proposed method when it is applied to training the model from scratch.
For fair comparisons, we evaluate the classification performance of Hi-Mapper trained with the \textit{full-training} scheme (350 epochs) and \textit{fine-tuning} scheme (baseline + 50 epochs) of the same learning objectives on ImageNet-1K~\cite{deng2009imagenet}.
As shown in~\tabref{tab:fvsft}, the experimental results demonstrate that the \textit{fine-tuning} scheme is better-suitable than \textit{full-training} in terms of understanding the structural organization of visual scenes.
\section{Additional visualization}\label{supp:addvis}
\begin{figure*}[t]
    \centering
    \includegraphics[width=\textwidth]{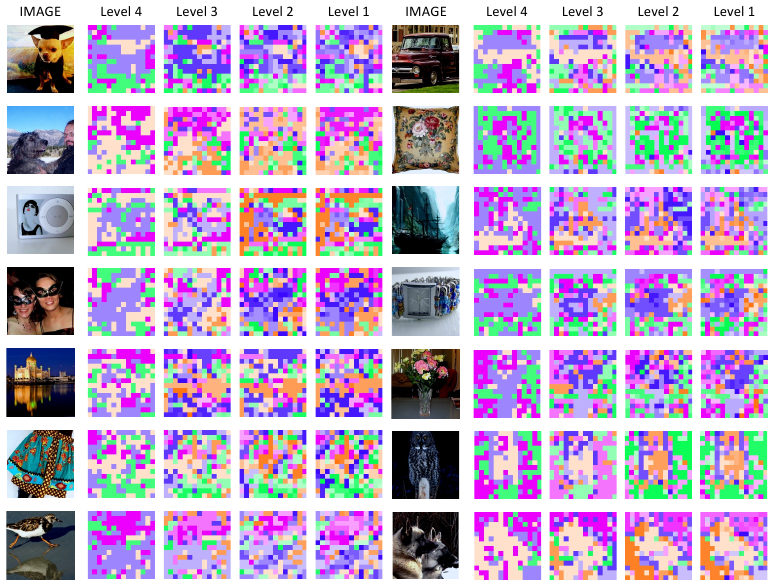}
    \vspace{-5pt}
    \caption{
 Visualization of visual hierarchy trees decomposed by Hi-Mapper(DeiT-S) trained on ImageNet-1K with classification objective. The same color family represents the same subtree.}
\vspace{-10pt}
\label{fig:suppl1}
\end{figure*}

\begin{figure*}[t]
    \centering
    \includegraphics[width=\textwidth]{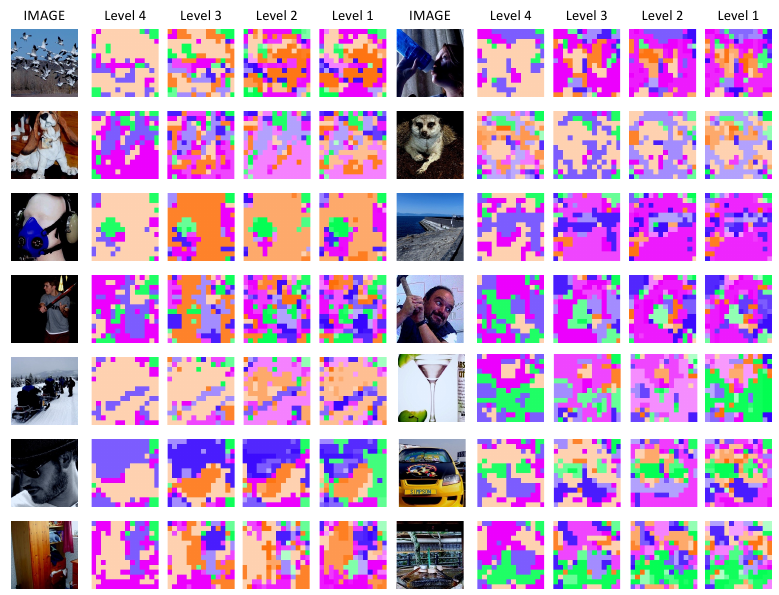}
    \vspace{-5pt}
    \caption{
 Visualization of visual hierarchy trees decomposed by Hi-Mapper(ENB4) trained on ImageNet-1K~\cite{deng2009imagenet} with classification objective. The same color family represents the same subtree.}
\vspace{-10pt}
\label{fig:suppl2}
\end{figure*}

For a more comprehensive understanding, we will provide additional visualization results that are included in the main paper and also examine the visual hierarchy in CNNs~\cite{efficientnet}, as shown in Figure \ref{fig:suppl1}, \ref{fig:suppl2}.  
This will offer insights into the feature representation aspects in transformer structures and CNNs, as well as the benefits of applying our method.


%% file: supple/fvsft.tex
\begin{table}[t]
\centering
\caption{Performance comparisons between \textit{full-trining} and \textit{fine-tuning} across various DNNs on the ImageNet-1K dataset~\cite{deng2009imagenet}.}
\label{tab:fvsft}
\small{
\begin{tblr}{
  width = \linewidth,
  colspec = {Q[240]Q[240]Q[240]Q[150]},
  column{even} = {c},
  column{3} = {c},
  vline{2,4} = {-}{},
  hline{1-2,7} = {-}{},
}
Backbone & \textit{full-training}   & \textit{fine-tuning}     & $\Delta$ \\
DeiT-T~\cite{deit}   & 74.5\%          & \textbf{74.8}\% & \blue{+0.3}                                                    \\
DeiT-S~\cite{deit}   & \textbf{82.8}\% & 82.6\%          & \red{-0.2}                                                    \\
DeiT-B~\cite{deit}   & 83.3\%          & \textbf{83.4}\% & \blue{+0.1}                                                    \\
Swin-T~\cite{swin}   & 83.2\%          & \textbf{83.5}\% & \blue{+0.3}                                                    \\
Swin-S~\cite{swin}   & 83.6\%          & \textbf{84.1}\% & \blue{+0.5}
\end{tblr}}
\end{table}